# A Multivariate Biomarker for Parkinson's Disease

Giancarlo Crocetti, Michael Coakley, Phil Dressner, Wanda Kellum, Tamba Lamin
Seidenberg School of Computer Science and Information Systems
Pace University
White Plains, NY, USA
{gcrocetti, mcoakley, pd50340n, tl98810w, wk59882w}@pace.edu

*Abstract*—In this study, we executed a genomic analysis with the objective of selecting a set of genes (possibly small) that would help in the detection and classification of samples from patients affected by Parkinson Disease. We performed a complete data analysis and during the exploratory phase, we selected a list of differentially expressed genes. Despite their association with the diseased state, we could not use them as a biomarker tool. Therefore, our research was extended to include a multivariate analysis approach resulting in the identification and selection of a group of 20 genes that showed a clear potential in detecting and correctly classify Parkinson Disease samples even in the presence of other neurodegenerative disorders.

## I. Introduction

We will analyze the microarray expression data of patients affected by Parkinson's disease (PD) with the goal of identifying a biomarker for this condition. The expression dataset used in this research is the Parkinson_105_from_CEL.xls file [1] containing the data from the GEO accession GSE6613 generated using the MAS5 algorithm.

The dataset contains a total of 22,283 measurements of gene expression from samples belonging to three distinct groups of people for a total of 105 samples:

1. Parkinson's disease group (50 patients)
2. Healthy control group. (22)
3. Neurodegenerative control group. (33)

The disease control group (3) contains samples from patients with various neurodegenerative diseases: from Alzheimer to system atrophy. This control group might have characteristics that mimic symptoms of PD and will help to increase the specificity of the resulting biomarker.

## II. Data Preparation

The sample's columns in the original dataset have been renamed in order to easily identify the class to which each sample belongs. In particular the name for each sample has been prefixed with HC, ND or PD where:

HC: Health Control sample
ND: Neurodegenerative sample
PD: Parkinson Disease sample

The prefix is followed by a sequence number for a unique identification within the same group.

Example: "HC_01_log_z" represents a healthy control samples that has been transformed and standardized.

Before any scaling and/or normalization on the data we removed probesets whose expression measurements are not reliable or represent experimental noise. As filtering method we used the "*filtering by Present calls*" with a threshold of 25%. Relaxing this threshold to a value of 25% is reasonable due to the high number of samples in the dataset as suggested in [2]:

"For data sets with 3-4 samples 50% Present spares most of the probe sets significant at $p \leq 0.001$ and those probes sets found most consistently. For more samples, relaxing the threshold to 25% fraction Present is reasonable".

After the filtering phase the number of probesets dropped from 23,283 to 8,100. During this process, we also eliminated probesets with expression amplitude below the noise level (<100) since they represent nothing but noise.

The data was, then, analyzed and as expected it showed a strong right-skewness. In figure 1 we can see the effect of the logarithmic transformation on one of the control sample.

FIGURE 1. EFFECT OF LOGARITHMIC TRANSFORMATION

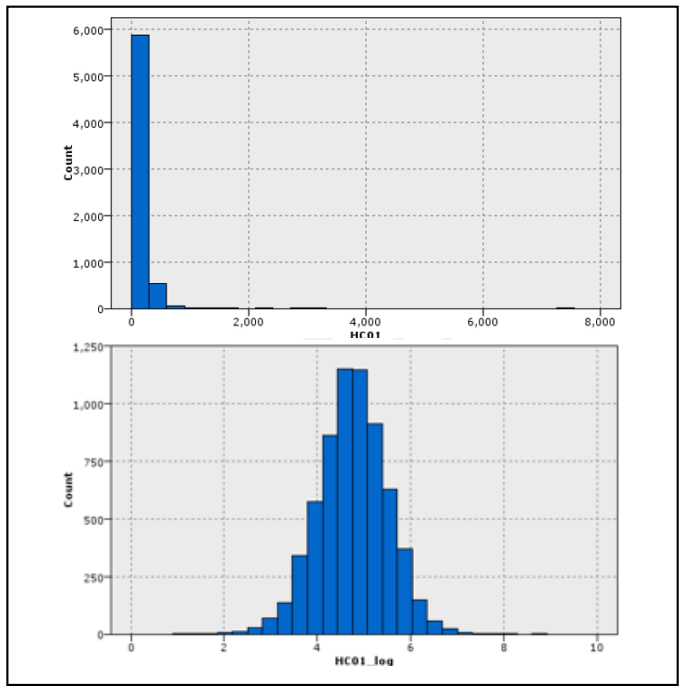

We then derived the z-score for each probeset defined as:

$$z_i = \frac{x_i - \mu_i}{\sigma_i}$$

where $\mu_i$ and $\sigma_i$ are the mean and standard deviation of the $i^{th}$ probeset expression levels.

We used the z-score for normalizing the data, but also to detect the presence of possible outliers. During this process we were able to find some anomalous values in the dataset. By anomalous value we mean an observation that appears to be inconsistent when compared with values belonging to the same class for the same gene. We marked as possible outlier any value with an absolute z score value greater than 5.

An example of possible outlier can be found in the 49th sample for the 200028_at probeset in the Parkinson class. In this case the expression level is too low when compared to the other measurement in the same class. In fact, the average expression value in this class is 215.216; however the expression value for this particular sample is 30.72 indicating a possible defective probe.

Since outliers might negatively influence analysis results we decided to investigate more closely and found a total of 643 outliers. How to deal with outliers in microarray data is not a straightforward topic, so we decided to replace these values with the average within their respective class: this might not be the best solution, however, it will limit the influence of such anomalous values during the data processing phase, especially in cases of algorithms sensitive to outliers.

As input in the next phases we prepared a dataset that:

- Does not contain unreliable or noise probesets.
- Does not contain outlier.
- Contains normalized and standardized value.

We decided on these characteristics by noticing that almost all algorithms have better performance with a dataset that has been normalized and standardized. By "better performance" we mean at least one of the following:

- An improvement on processing time.
- An improvement on classification.

### III. EXPLORATORY DATA ANALYSIS

We performed the univariate analysis with the goal of building a ranking of differentially expressed genes. We point out that this result is a univariate result in which each gene is considered in isolation and excluding its (possibly important) interaction with all other genes in the dataset; consequently the list of genes resulting from this task should not be used as biomarker in a classification system.

The problem of building a biomarker for Parkinson disease will be considered in the next sections.

For the basic exploratory analysis we process our data using the "GenePattern" application from the MIT institute [3].

For each gene, the application uses a test statistic to calculate the difference in gene expression between classes and then computes a p-value to estimate the significance of the test statistic score. Because testing tens of thousands of genes simultaneously increases the possibility of mistakenly identifying a non-differentially expressed gene as a differentially expressed gene (a false positive), GenePattern corrects for multiple hypothesis testing by computing both false discovery rates (FDR) and family-wise error rates (FWER) adjusted using the Bonferroni/Hochberg method. Researchers using GenePattern generally identify differentially expressed genes based on FDR rather than the more conservative FWER.

The first list of differentially expressed genes was obtained by running the dataset using the adjusted ANOVA F-test statistic with 1,000 permutations on all three sample classes. In order to compare results we ran the same test using the SNR (Signal to Noise Ratio) and the results were identical showing that, for this dataset, the adjusted ANOVA F-test and SNR results coincided when compared to the final ranking of genes.

The test statistics are defined in GenePattern as the following:

t-Test: $\dfrac{\mu_A - \mu_B}{\sqrt{\dfrac{\sigma_A^2}{n_A} + \dfrac{\sigma_B^2}{n_B}}}$

where μ_A, μ_b are the group averages, δ_A, δ_B the group standard deviations, and $n_A$, $n_B$ the group sizes.

SNR: $\dfrac{\mu_A - \mu_B}{\sigma_A + \sigma_B}$

where μ_A, μ_B are the group averages and δ_A, δ_B the group standard deviations

The application of this technique on a multiclass dataset generated a list of genes that:

- Are up-regulated in PD (Parkinson Disease) when compared with the other classes
- Are up-regulated in the other classes but down-regulated in PD. The result of this analysis does not indicate which class contains the up-regulated gene(s), so a further analysis is warrant between the healthy control and the neurodegenerative samples.

FIGURE 2. UP REGULATED FEATURES

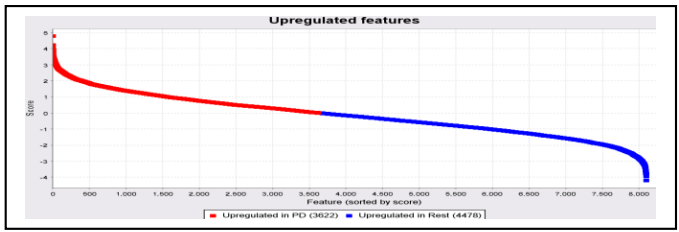

From the graph we were able to identify 60 genes with an adjusted FDR (Bonferroni/Hochberg) less than 0.01.

We conclude this section by displaying the heat map of the top 40 genes as in figure 3. It was our intention to display the heat map for all 60 genes, but it would have been impossible to fit in this page.

FIGURE 3. HEAT MAP FOR THE TOP 40 GENES RANKED BY FDR

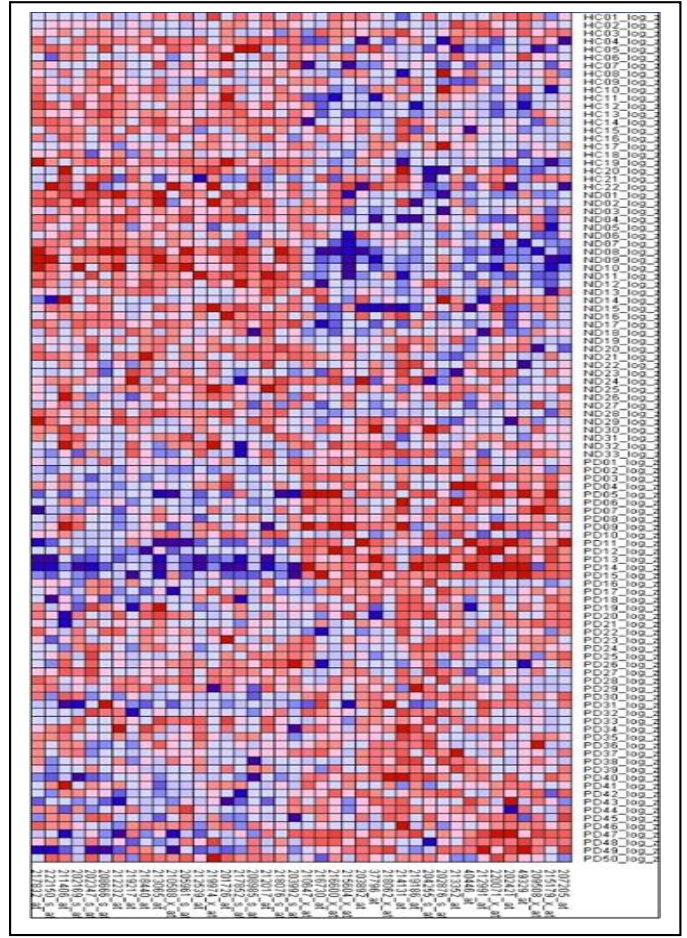

In the image we can clearly distinguish areas in which the genes in each class are up or down regulated. The difference is more evident between samples in the Parkinson class when compared with the other two. More importantly we can clearly distinguish the difference in expression levels of genes between the Parkinson and the neuro-degenerative classes. This might be an important finding since the neuro-degenerative class has been introduced in the dataset in order to verify the prediction power of any biomarker for Parkison disease.

## IV. MULTIVARIATE MODELLING

In this section we will try to build a subset of features (as small as possible) that will be used as a multivariate classification model that will differentiate classes represented in the data set.

Multivariate analysis considers the simultaneous effect of genes instead of stopping to the influence of single genes, which is typical of univariate approaches [4].

For this particular task we used the feature selection of Weka [5] ver. 3.4 called "Attribute Selection".

Weka provides a rich set of multivariate algorithms and we considered:

- *Wrapper Subset Evaluator (WSE)*: implementation of forward wrapper method for feature selection for the creation of an optimal subset.
- *Correlation-based Feature Selection* (CFS): evaluation of different combinations of features to identify an optimal subset. The feature subsets are generated using different search techniques. We used Best First and Greedy search methods with a forward direction.
- *R-Support Vector Machine (RSVM)*: SVM in its recursive version.

TABLE 1 – GENES SETS IDENTIFIED BY EACH MODEL

| Algorithms | # of features Selected | Selected Subset |
|---|---|---|
| WSE | 6 | 200639_s_at 202690_s_at 203303_at 207730_x_at 211275_s_at 217301_x_at |
| SVM | 20 | 202581_at 214800_x_at 208843_s_at 220897_at 207205_at 212994_at 219055_at 220471_s_at 212176_at 204031_s_at 201186_at 219156_at 219186_at 217142_at 206342_x_at 33814_at 213891_s_at 213340_s_at 217552_x_at 211989_at |
| CFS | 39 | 200994_at 201935_s_at 202169_s_at 202213_s_at 202258_s_at 202347_s_at 202690_s_at 202727_s_at 202778_s_at 203104_at 203116_s_at 203153_at 203273_s_at 203303_at 203992_s_at 204255_s_at 207205_at 207416_s_at 208666_s_at 209048_s_at 209303_at 210647_x_at 210858_x_at 211406_at 213596_at 214800_x_at 215158_s_at 216341_s_at 216524_x_at 216600_x_at 217301_x_at 217819_at 217922_at 218236_s_at 218680_x_at 219055_at 220529_at 221192_x_at AFFX-HSAC07/X00351_M_at |

In all cases, we used 10 folds cross-validation method during the feature selection process, and in table 2 we report the parameters used in configuring the various algorithms in order to produce the feature subset.

TABLE 2 – MODEL CONFIGURATION

| Algorithm | Implementation | Parameters |
|---|---|---|
| WSE | **WrapperSubsetEval** | **Rules**: DecisionTable **Fold**: 5 **Search**: GreedyStepwise |
| SVM | **SVMAttributeEval** | **Attribute to Eliminate per Iteration**: 1 **Filter Type**: No normalization **Search**: Ranker |
| CFS | **CfsSubsetEval** | **Search**: GreedyStepwise |

Any parameter not expressively present in the table has been used with its default value.

In order to validate each model we used the "Weka Flow Knowledge" environment that allows the creation of cross-validation set and verify the classification model as shown in figure 4.

FIGURE 4. WEKA KNOWLEDGE FLOW

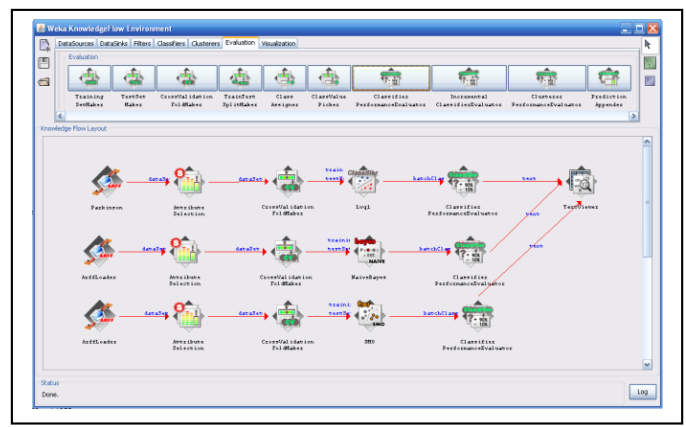

The "Cross Validation Fold Maker" performs a cross-validation, in which feature selection is performed during training of each model.

We noticed that each model performs better with its own particular classifier: so WSE performs better with NaiveBayes, CFS achieves high classification rates using neural networks LVQ while SVM performs well with SMO.

TABLE 3 – CLASSIFICATION RESULTS

| Model | Classification Results |
|---|---|
| WSE | ```
= Stratified cross-validation = Summary =

Correctly Classified     (69)       65.7143 %
Incorrectly Classified (36)         34.2857 %
Kappa statistic                      0.4011
Mean absolute error                  0.3245
Root mean squared error              0.4182
Relative absolute error             77.0204 %
Root relative squared error         91.1559 %
Total Number of Instances          105

=== Confusion Matrix ===
  a  b  c   <-- classified as
  4  6 12 |  a = HC
  0 17 16 |  b = ND
  0  2 48 |  c = PD
``` |
| CFS | ```
= Stratified cross-validation = Summary =

Correctly Classified     (59)       73.75  %
Incorrectly Classified (21)         26.25  %
Kappa statistic                      0.577
Mean absolute error                  0.175
Root mean squared error              0.418
Relative absolute error             41.403 %
Root relative squared error         91.013 %
Total Number of Instances          105

=== Confusion Matrix ===

  a  b  c   <-- classified as
  9  3  5 |  a = HC
  1 19  5 |  b = ND
  3  4 31 |  c = PD
``` |
| RSVM | ```
= Stratified cross-validation = Summary =

Correctly Classified     (87)       82.8571 %
Incorrectly Classified (18)         17.1429 %
Kappa statistic                      0.7228
Mean absolute error                  0.2667
Root mean squared error              0.3432
Relative absolute error             63.2941 %
Root relative squared error         74.8253 %
Total Number of Instances          105

=== Confusion Matrix ===

  a  b  c   <-- classified as
 16  0  6 |  a = HC
  2 26  5 |  b = ND
  1  4 45 |  c = PD
``` |

From the results in table 3, we can see how the RSVM algorithm performed better than WSE and CFS even though WSE provided the smallest subset with only 6 genes.

The higher classification rate for the SVM is due to its capability of correctly identify the difference between ND and PD, something that is missing from WSE and CFS which incorrectly misclassified many healthy tissues.

The many algorithms had many options and tuning parameters and we are confident that all three classification models could be improved both at the feature selection and classification levels.

## V. CONCLUSIONS AND FURTHER WORK

Multivariate models are a necessary tools in genomic studies that could be applied for the detection and identification of neurodegenerative conditions like Parkinson's disease.

Among the algorithms tested in this study, RSVM clearly came out as an effective model to adopt in biomarker discovery, with the important ability of successfully discriminate between PD and other neurodegenerative diseases.

The identification of a small multivariate set of genes associated to PD is an important step forward in Genomics, but this research cannot stop here, and the natural next step is to look for the biological interpretation of this result.